\documentclass{article}

\usepackage{arxiv}

\usepackage[utf8]{inputenc} 
\usepackage[T1]{fontenc}    
\usepackage{hyperref}       
\usepackage{amsmath, amssymb}

\usepackage[linesnumbered, ruled, vlined]{algorithm2e}
\usepackage{bm}
\usepackage{subcaption}
\usepackage{booktabs}   
\usepackage{multirow}   
\usepackage{ulem}       
\usepackage{siunitx}    
\usepackage{caption}    
\usepackage{fontawesome5}

\usepackage{url}            
\usepackage{booktabs}       
\usepackage{amsfonts}       
\usepackage{nicefrac}       
\usepackage{microtype}      
\usepackage{cleveref}       
\usepackage{lipsum}         
\usepackage{graphicx}
\usepackage{natbib}
\usepackage{doi}

\title{One-Step Diffusion with Inverse Residual Fields for Unsupervised Industrial Anomaly Detection}

\newif\ifuniqueAffiliation
\uniqueAffiliationtrue

\ifuniqueAffiliation 
\usepackage{fontawesome5}

\newcommand{\corrauthor}{%
  \textsuperscript{\smash{\rlap{\hspace{0.2em}\faEnvelope}}}%
}

\newcommand{\AuthorBlock}[5]{%
  \parbox[t]{0.44\textwidth}{%
    \centering
    \small
    \textbf{#1}\strut\\
    #2\strut\\
    #3\strut\\
    #4\strut\\
    #5\strut
  }%
}

\newcommand{\CST}{Department of Computer Science and Technology}
\newcommand{\EE}{Department of Electronic Engineering}

\author{
  \AuthorBlock
    {Boan Zhang}
    {\CST}
    {Xidian University}
    {Shaanxi, China 710126}
    {\mbox{}}
  \And
  \AuthorBlock
    {Wen Li}
    {\CST}
    {Xidian University}
    {Shaanxi, China 710125}
    {\mbox{}}
  \AND
  \AuthorBlock
    {Guanhua Yu}
    {\CST}
    {Xidian University}
    {Shaanxi, China 710126}
    {\mbox{}}
  \And
  \AuthorBlock
    {Xiyang Liu}
    {\CST}
    {Xidian University}
    {Shaanxi, China 710125}
    {\mbox{}}
  \AND
  \AuthorBlock
    {Wenchao Chen\corrauthor}
    {\EE}
    {Xidian University}
    {Shaanxi, China 710126}
    {\texttt{chenwenchao@xidian.edu.cn}}
  \And
  \AuthorBlock
    {Long Tian\corrauthor}
    {\CST}
    {Xidian University}
    {Shaanxi, China 710126}
    {\texttt{tianlong@xidian.edu.cn}}
}
\else
\usepackage{authblk}

\newbox{\orcid}\sbox{\orcid}{\includegraphics[scale=0.06]{orcid.pdf}} 
\author[1]{%
	\href{https://orcid.org/0000-0000-0000-0000}{\usebox{\orcid}\hspace{1mm}David S.~Hippocampus\thanks{\texttt{hippo@cs.cranberry-lemon.edu}}}%
}
\author[1,2]{%
	\href{https://orcid.org/0000-0000-0000-0000}{\usebox{\orcid}\hspace{1mm}Elias D.~Striatum\thanks{\texttt{stariate@ee.mount-sheikh.edu}}}%
}
\affil[1]{Department of Computer Science, Cranberry-Lemon University, Pittsburgh, PA 15213}
\affil[2]{Department of Electrical Engineering, Mount-Sheikh University, Santa Narimana, Levand}
\fi

\renewcommand{\shorttitle}{\textit{arXiv} Template}

\hypersetup{
pdftitle={A template for the arxiv style},
pdfsubject={q-bio.NC, q-bio.QM},
pdfauthor={David S.~Hippocampus, Elias D.~Striatum},
pdfkeywords={First keyword, Second keyword, More},
}

\begin{document}
\maketitle

\begin{abstract}
	Diffusion models have achieved outstanding performance in unsupervised industrial anomaly detection (uIAD) by learning a manifold of normal data under the common assumption that off-manifold anomalies are harder to generate, resulting in larger reconstruction errors in data space or lower probability densities in the tractable latent space. However, their iterative denoising and noising nature leads to slow inference. In this paper, we propose OSD-IRF, a novel one-step diffusion with inverse residual fields, to address this limitation for uIAD task. We first train a deep diffusion probabilistic model (DDPM) on normal data without any conditioning. Then, for a test sample, we predict its inverse residual fields (IRF) based on the noise estimated by the well-trained parametric noise function of the DDPM. Finally, uIAD is performed by evaluating the probability density of the IRF under a Gaussian distribution and comparing it with a threshold. Our key observation is that anomalies become distinguishable in this IRF space, a finding that has seldom been reported in prior works. Moreover, OSD-IRF requires only single step diffusion for uIAD, thanks to the property that IRF holds for any neighboring time step in the denoising process. Extensive experiments on three widely used uIAD benchmarks show that our model achieves SOTA or competitive performance across six metrics, along with roughly a 2$\times$ inference speedup without distillation.
\end{abstract}


\section{Introduction}
\label{sec:introduction}
Industrial anomaly detection (IAD) plays a crucial role in applications such as industrial quality inspection and product safety \cite{pang2021deep}. Its goal is to automatically identify industrial images that deviate from normal patterns. In real-world production lines, anomalous images are scarce due to the high cost of acquisition, making supervised IAD methods challenging to apply \cite{kamat2020anomaly}. Consequently, unsupervised IAD (uIAD) methods, trained on large amounts of normal image data, have been developed \cite{liu2023simplenet, roth2022towards}.
The success of uIAD heavily relies on learning a compact high-dimensional distribution of normal data, from which anomalous images become distinguishable at test time. Diffusion-based generative models (DGMs) have therefore emerged as powerful tools for uIAD \cite{wyatt2022anoddpm} due to their robust capability to handle high-dimensional data and approximate complex distributions \cite{ho2020denoising, song2020score, song2020denoising}.
DGMs for uIAD can be mainly divided into two lines of works, reconstruction-based methods (e.g., OmiAD \cite{feng2025omiad}) and density estimation-based methods (e.g., InvAD \cite{sakai2025invad}).

\begin{figure}[t]
	\centering
	\centerline{\includegraphics[width=16cm]{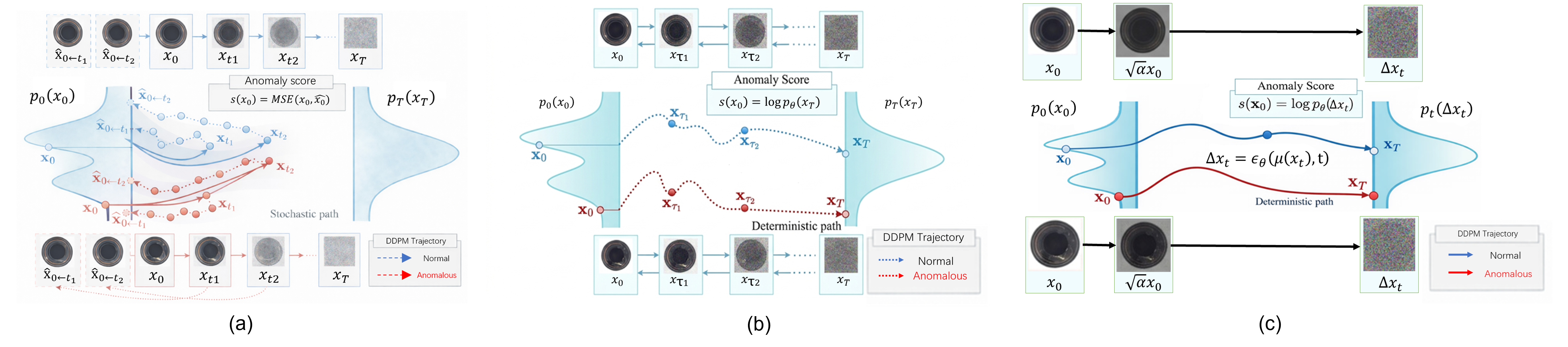}}
	\vspace{-2mm}
	\caption{
		Graphical comparisons of the previous methods and our proposed OSD-IRF. (a) Illustration of a previous reconstruction-based diffusion model, which conducts anomaly detection (AD) in the input data space $\boldsymbol{x}_0$. It first perturbs an image $\boldsymbol{x}_0$ to a latent state $\boldsymbol{x}_t$ using a noise function $\boldsymbol{\epsilon}$, then denoises $\boldsymbol{x}_t$ back to $\hat{\boldsymbol{x}}_0$ using a well-trained parametric noise function $\boldsymbol{\epsilon}_{\boldsymbol{\theta}}(\cdot, t)$. The anomaly score is computed as the mean squared error (MSE) between $\boldsymbol{x}_0$ and $\hat{\boldsymbol{x}}_0$. (b) Illustration of a previous density estimation-based diffusion model, which performs AD in the noisy latent state space $\boldsymbol{x}_T$. It uses a well-trained parametric noise function $\boldsymbol{\epsilon}_{\boldsymbol{\theta}}(\cdot, t)$ to perturb an image $\boldsymbol{x}_0$ to a noisy latent state $\boldsymbol{x}_T$, where $\boldsymbol{x}_T$ follows a standard Gaussian distribution. The anomaly score is determined by the typicality of $\boldsymbol{x}_T$ within this tractable distribution. (c) Unlike the previous methods in (a) and (b), our proposed OSD-IRF carries out AD in the inverse residual fields (IRF) space $\Delta \boldsymbol{x}_t= \frac{1}{\sqrt{\beta_t}}(\boldsymbol{x}_t - \sqrt{1-\beta_t} \boldsymbol{x}_{t-1})$.
		The anomaly score is then determined based on the typicality of $\Delta \boldsymbol{x}_t$ under a Gaussian distribution. Notably, the computation of $\Delta \boldsymbol{x}_t$ does not require multi-step inference (See Sec.\ref{sec:irf}), our model not only achieves SOTA performance but also reduces the inference to a single step.}
	\label{fig:motivation}
	\vspace{-2mm}
\end{figure}

Reconstruction-based methods typically treat anomalies as noise and leverage a diffusion model trained exclusively on normal images to denoise them. As a result, anomalous features can be effectively removed and reconstructed into normal ones. Anomalies are then detected based on the discrepancy between the input images (or features) and their reconstructed counterparts, as illustrated in Fig. \ref{fig:motivation} (a).
\cite{lu2023removing} introduces random noise to overwhelm anomalous pixels and obtains pixel-wise anomaly scores from the intermediate denoising process.
\cite{he2024diffusion, mousakhan2024anomaly} simultaneously propose using the input image as a condition to guide the denoising process, resulting in a defect-free reconstruction while preserving nominal patterns.
Different from Gaussian diffusion models, \cite{wyatt2022anoddpm} develops a multi-scale simplex noise diffusion process that offers control over the target anomaly size.
To speed up inference, OmiAD \cite{feng2025omiad} introduces a one-step denoising model derived from a well-designed multi-step adaptive masked diffusion model and compressed via adversarial score distillation. Despite its effectiveness, OmiAD is difficult to train due to the instability of adversarial learning.

Density estimation-based methods leverage the denoising process, which implicitly learns the score function of the data distribution, to enable likelihood-based anomaly detection, as illustrated in Fig. \ref{fig:motivation} (b).
DTE \cite{livernoche2023diffusion} estimates the distribution over diffusion time for a given input image and uses the mean of this distribution as the anomaly score.
InvAD \cite{sakai2025invad} directly infers the final latent variable corresponding to the input image via denoising diffusion implicit model (DDIM) \cite{song2020denoising} inversion and measures the deviation based on the known Gaussian distribution for anomaly scoring.
However, DTE requires iterating over all reverse time steps to compute the anomaly score, while InvAD's anomaly score relies on using the well-trained parametric noise function of the denoising model to transform the input data into a Gaussian distribution over multiple time steps.

To address the aforementioned issues in reconstruction-based and density estimation-based methods, this paper proposes OSD-IRF, a one-step diffusion with inverse residual fields for uIAD.  It belongs to the density estimation-based paradigm. However, unlike previous work in this category, our core insight is that anomalies become distinguishable in the inverse residual fields (IRF), a property that has rarely been explored, as illustrated in Fig. \ref{fig:motivation} (c).
Specifically, we first train a DDPM \cite{ho2020denoising} on normal image data. At test time, we define the IRF as the predicted noise when transitioning $\boldsymbol{x}_t$ from time step $t$ to $t-1$ with the well-trained noise function of DDPM, denoted by $\Delta \boldsymbol{x}_t= \frac{1}{\sqrt{\beta_t}}(\boldsymbol{x}_t - \sqrt{1-\beta_t} \boldsymbol{x}_{t-1})$.
The anomaly score is then obtained by computing the log-likelihood of the predicted IRF $\Delta \boldsymbol{x}_t$ under a standard Gaussian distribution.
The novelty and contributions of our proposed OSD-IRF can be summarized as follows:

\noindent (1) OSD-IRF performs anomaly detection in IRF space defined by the predicted noise from time-step $t$ to $t-1$. Its one-step diffusion property enables both powerful distribution modeling and fast inference for uIAD task.

\noindent (2) OSD-IRF is a plug-and-play inference technique compatible with any diffusion model, regardless of architecture or training objective. Without loss of generality, we adopt DDPM as our backbone in this paper.

\noindent (3) OSD-IRF achieves SOTA or competitive performance on three widely used uIAD benchmarks across six popular metrics, along with roughly 2 $\times$ inference speedup without distillation.

\section{Preliminaries}
\label{sec:preliminaries}
\subsection{Diffusion Model}
DDPM \cite{ho2020denoising} is a prominent example of diffusion models \cite{song2020score, rombach2022high}, which comprises a forward process along with a reverse denoising process. In the forward process, noise is incrementally introduced, ultimatedly coverting the initial image data variable $\boldsymbol{x}_0$ into Gaussian noise $\boldsymbol{x}_T$ over $T$ time steps:
\begin{align}
	 & q(\boldsymbol{x}_{1:T} | \boldsymbol{x}_0) = \prod_{t=1}^{T} q(\boldsymbol{x}_t|\boldsymbol{x}_{t-1}), \quad q(\boldsymbol{x}_t|\boldsymbol{x}_{t-1}) = \mathcal{N}(\sqrt{1-\beta_t} \boldsymbol{x}_{t-1}, \beta_t \textbf{I}) \label{eq:ddpm1}
\end{align}
where $\beta_t$ is the noise level that typically set to be a small constant. A notable characteristic of the forward process descriping the relationship between $\boldsymbol{x}_t$ and $\boldsymbol{x}_0$ can be derived recursively accroding to Eq. \ref{eq:ddpm1} as:
\begin{align}
	q(\boldsymbol{x}_t|\boldsymbol{x}_0) = \mathcal{N}(\boldsymbol{x}_t; \sqrt{\alpha_t} \boldsymbol{x}_0, (1-\alpha_t) \textbf{I}) \label{eq:ddpm2}
\end{align}
where $\alpha_t = \prod_{t=1}^T (1-\beta_t)$.
Utilizing a Markov chain with trainable Gaussian transitions, the denoising process from $\boldsymbol{x}_t$ back to $\boldsymbol{x}_0$ unfolds as:
\begin{align}
	 & p_{\boldsymbol{\theta}}(\boldsymbol{x}_{0:T}) = p_{\boldsymbol{\theta}}(\boldsymbol{x}_T) \prod_{t=1}^T p_{\boldsymbol{\theta}}(\boldsymbol{x}_{t-1}|\boldsymbol{x}_t), \quad p_{\boldsymbol{\theta}}(\boldsymbol{x}_{t-1}|\boldsymbol{x}_t) = \mathcal{N}(\boldsymbol{\mu}_{\boldsymbol{\theta}}(\boldsymbol{x}_t,t), \sigma_t^2 \textbf{I}) \label{eq:ddpm3}
\end{align}
where $\boldsymbol{\mu}_{\boldsymbol{\theta}}(\boldsymbol{x}_t,t) = \frac{1}{\sqrt{\alpha_t}}(\boldsymbol{x}_t - \frac{\beta_t}{\sqrt{1-\alpha_t}}\boldsymbol{\epsilon}_{\boldsymbol{\theta}}(\boldsymbol{x}_t, t))$.
The forward process provides ground-truth signals for training the parameters of the noise prediction network $\boldsymbol{\epsilon}_{\boldsymbol{\theta}}(\cdot)$ in the denoising process. Following the derivations in \cite{ho2020denoising, han2022card}, $\boldsymbol{\epsilon}_{\boldsymbol{\theta}}(\cdot)$ is trained to minimize the regression loss defined as:
\begin{align}
	\mathcal{L} = {\rm{min}}_{\boldsymbol{\theta}} \mathbb{E}_{t,\boldsymbol{x}_0,\boldsymbol{\epsilon} \sim \mathcal{N}(\textbf{0},\textbf{I})} \| \boldsymbol{\epsilon} - \boldsymbol{\epsilon}_{\boldsymbol{\theta}}(\boldsymbol{x}_t, t) \|_2^2 \label{eq:ddpm4}
\end{align}
In this paper, we utilize powerful generative capability of the well-trained DDPM to obtain distinguishable inverse residual fields (IRF, see details in Sec. \ref{sec:irf}) for the robust and fast uIAD.

\subsection{Diffusion-based uIAD}
As illustrated in Fig. \ref{fig:motivation} (a) and (b), there are mainly two types of diffusion-based uIAD methods. We first state the assumptions underlying how uIAD can be conducted with these two lines of work, and then briefly introduce their pipelines for implementing uIAD.

Reconstruction-based models \cite{wyatt2022anoddpm} assume that normal images lie within a normal manifold and can be well reconstructed, while abnormal patterns deviate from this manifold and thus cannot be faithfully reconstructed. The anomaly score is then derived from the discrepancy between the input image (or features) and its reconstructed counterpart. In practice, such a model first perturbs an image $\boldsymbol{x}_0$ to a latent state $\boldsymbol{x}_t$ using a noise function $\boldsymbol{\epsilon} \sim \mathcal{N}(\textbf{0}, \textbf{I})$ following the reparameterized form \cite{kingma2013auto} of Eq. \ref{eq:ddpm2} as:
\begin{align}
	\boldsymbol{x}_t = \sqrt{1-\beta_t} \boldsymbol{x}_{t-1} + \sqrt{\beta_t} \boldsymbol{\epsilon} \label{eq:ddpm5}
\end{align}
Then it denoises $\boldsymbol{x}_t$ back to $\hat{\boldsymbol{x}}_0$ over multiple time steps using a well-trained parametric noise function $\boldsymbol{\epsilon}_{\boldsymbol{\theta}}(\cdot, t)$ from Eq. \ref{eq:ddpm4}. Additionally, conditioning information $\boldsymbol{c}$ can be incorporated by merging it with $\boldsymbol{x}_t$ \cite{he2024diffusion, mousakhan2024anomaly} to further improve uIAD performance.

Density estimation-based models \cite{sakai2025invad} assume the existence of a tractable distribution over certain latent states, under which normal and abnormal patterns are probabilistically distinguishable. For example, normal patterns lie in high-probability regions, while abnormal patterns fall into low-probability regions. In practice, such a model directly uses a well-trained parametric noise function $\boldsymbol{\epsilon}_{\boldsymbol{\theta}}(\cdot, t)$ from Eq. \ref{eq:ddpm4} to perturb an image $\boldsymbol{x}_0$ to a latent state $\boldsymbol{x}_T$ (typically $T \geq t$) over multiple time steps using jump pertubation (e.g., DDIM \cite{song2020denoising}) for speedup, where $\boldsymbol{x}_T \sim \mathcal{N}(\textbf{0}, \textbf{I})$. The anomaly score is then determined based on the typicality of $\boldsymbol{x}_T$ within the Gaussian distribution.

\section{Method}
\label{sec:method}
To address the limitations of current diffusion-based uIAD methods, we build upon InvAD \cite{sakai2025invad}, a most recently proposed density estimation-based model. The success of InvAD relies on two crucial components: a latent state with tractable distribution for calculating anomaly score, and a flexible function to derive that distribution from an input image data. InvAD employs the latent state $\boldsymbol{x}_T$ of the input image data at the $T$-th time step, which follows a Gaussian distribution, as the tractable distribution, and it uses the well-trained parametric noise function $\boldsymbol{\epsilon}_{\boldsymbol{\theta}}(\cdot, t)$ as the means to derive the tractable distribution from $\boldsymbol{x}_0$ to $\boldsymbol{x}_T$ in an iterative manner. Go one step further, we notice that using latent state $\Delta \boldsymbol{x}_t= \frac{1}{\sqrt{\beta_t}}(\boldsymbol{x}_t - \sqrt{1-\beta_t} \boldsymbol{x}_{t-1})$, termed inverse residual fields (IRF) in this paper, also yields a Gaussian distribution when derived using the same well-trained parametric noise function $\boldsymbol{\epsilon}_{\boldsymbol{\theta}}(\cdot, t)$. Importantly, thanks to the property that IRF holds at any neighboring time step, thus adopting $\Delta \boldsymbol{x}_t$ as the latent state for uIAD not only achieves SOTA or competitive performance but also accelerates inference significantly compared to InvAD.

\subsection{OSD-IRF for Anomaly Detection}
\label{sec:irf}
We introduce one-step diffusion with inverse residual fields (OSD-IRF) for uIAD task. We first propose inverse residual fields (IRF), from which one-step diffusion for uIAD naturally follows. The IRF is obtained using a parametric noise function $\boldsymbol{\epsilon}_{\boldsymbol{\theta}}(\cdot, t)$ from a DDPM pre-trained on normal images via Eq. \ref{eq:ddpm4}. Formally, according to Eq. \ref{eq:ddpm4} and \ref{eq:ddpm5},
the IRF $\Delta \boldsymbol{x}_t= \frac{1}{\sqrt{\beta_t}}(\boldsymbol{x}_t - \sqrt{1-\beta_t} \boldsymbol{x}_{t-1})$ can be expressed as:
\begin{align}
	 & \Delta \boldsymbol{x}_t  = \boldsymbol{\epsilon}_{\boldsymbol{\theta}}(\boldsymbol{x}_t, t), \quad \boldsymbol{x}_t = \sqrt{\alpha_t} \boldsymbol{x}_0 + \sqrt{1-\alpha_t} \boldsymbol{\epsilon} \label{eq:irf1}
\end{align}
where $\boldsymbol{\epsilon} \sim \mathcal{N}(\textbf{0}, \textbf{I})$. According to the constraint of Eq. \ref{eq:ddpm4}, $\boldsymbol{\epsilon}_{\boldsymbol{\theta}}(\boldsymbol{x}_t, t)$ is also expected to follow a Gaussian distribution, namely $\boldsymbol{\epsilon}_{\boldsymbol{\theta}}(\boldsymbol{x}_t, t) \sim \mathcal{N}(\textbf{0}, \textbf{I})$. We use the features extracted by a pre-trained backbone such as EfficientNet \cite{tan2019efficientnet} to serve as $\boldsymbol{x}_0$, since features exhibit invariance to low-level variations like noise or illumination, which enhances uIAD performance.
In practice, we find that using the mean of $\boldsymbol{x}_t$ for uIAD is more robust than directly using samples from Eq. \ref{eq:irf1}. Following the reparameterized trick in \cite{kingma2013auto}, in Eq. \ref{eq:irf1}, the mean of $\boldsymbol{x}_t$ is $\mu(\boldsymbol{x}_t)=\sqrt{\alpha_t} \boldsymbol{x}_0$. Finally, our proposed IRF $\Delta \boldsymbol{x}_t$ used in this paper can be expressed as:
\begin{align}
	 & \Delta \boldsymbol{x}_t  = \boldsymbol{\epsilon}_{\boldsymbol{\theta}}(\mu(\boldsymbol{x}_t), t), \quad \mu(\boldsymbol{x}_t)=\sqrt{\alpha_t} \boldsymbol{x}_0 \label{eq:irf2}
\end{align}
Obviously, the IRF in Eq. \ref{eq:irf2} reflects the predicted noise when transitioning $\boldsymbol{x}_t$ from time step $t$ to $t-1$. Moreover, since the well-trained parametric noise function $\boldsymbol{\epsilon}_{\boldsymbol{\theta}}(\cdot, t)$ is defined in the reverse process of the diffusion model, we refer to these residual fields as inverse residual fields.
To distinguish between the IRFs defined in Eq. \ref{eq:irf1} and Eq. \ref{eq:irf2}, we refer to the IRF in Eq. \ref{eq:irf1} as the IRF under noisy-state input, and that in Eq. \ref{eq:irf2} as the IRF under mean-path input. Unless otherwise specified, the term IRF hereafter refers to the IRF under mean-path input defined in Eq \ref{eq:irf2}.

\textbf{Remark}: According to the above definition, the IRF is particularly well-suited for the uIAD due to three valuable properties. Firstly, anomalies are easily distinguished using the IRF, as normal and abnormal patterns are probabilistically distinguishable within the tractable latent distribution. We provide a toy data analysis in Sec. \ref{sec:toydata} to further illustrate this point. Secondly, the discriminative capability of the IRF is easy to derive by training a standard DDPM without any carefully designed conditons (See Sec. \ref{sec:benefits} for experimental evidence). Thirdly, the IRF enables fast inference because computing Eq. \ref{eq:irf2} does not require multiple time steps, thereby achieving single step anomaly detection.

\subsection{Anomaly Score Calculation}
\label{sec:irfuiad}
Following InvAD \cite{sakai2025invad}, we compute score map $\boldsymbol{A} \in \mathbb{R}^{H \times W}$ as:
\begin{align}
	\hat{\boldsymbol{A}}[i,j] = \|\Delta \boldsymbol{x}_t\|_2 = \sqrt{\sum_{r=1}^c \Delta \boldsymbol{x}_t^2[r,i,j]}, \quad \boldsymbol{A} = {\rm{Upsample}}(\hat{\boldsymbol{A}}) \label{eq:irf3}
\end{align}
where ${\rm{Upsample}}(\cdot)$ denotes bilinear interpolation.
Hence, the pixel-level anomaly detection (anomaly localization) is directly obtained from the score map $\boldsymbol{A}$. For (image-level) anomay detection, the anomaly score is computed as:
\begin{align}
	 & s = s_{\rm{diff}} + s_{\rm{nll}}, \label{eq:irf4}                                                                            \\
	 & s_{\rm{diff}} = {\rm{max}}_{(i,j)} \hat{\boldsymbol{A}}[i,j] - {\rm{min}}_{(i,j)} \hat{\boldsymbol{A}}[i,j], \label{eq:irf5} \\
	 & s_{\rm{nll}} = -{\rm{log}} \{\sum_{r,i,j}-\frac{1}{2}(\Delta \boldsymbol{x}_t^2[r,i,j])\} \label{eq:irf6}
\end{align}
where $s_{\rm{diff}}$ is the difference between the maximum and minimum values of the feature-scale score map $\hat{\boldsymbol{A}}$ in Eq. \ref{eq:irf4}, and $s_{\rm{nll}}$ is derived from the negtive log-likelihood of the IRF $\Delta \hat{\boldsymbol{x}}_t$ under a standard Gaussian distribution.

\begin{figure}[t]
	\centering
	\begin{subfigure}[t]{0.46\textwidth}
		\centering
		\includegraphics[width=\textwidth]{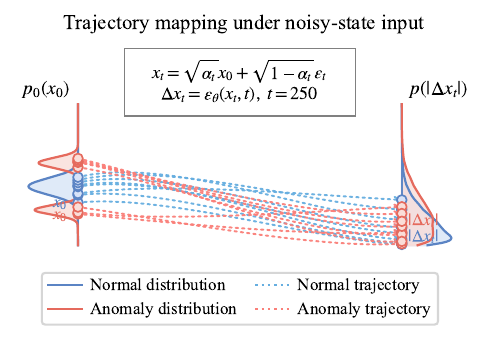}  
		\caption{}  
		\label{fig:2a}
	\end{subfigure}
	\hfill
	\begin{subfigure}[t]{0.46\textwidth}
		\centering
		\includegraphics[width=\textwidth]{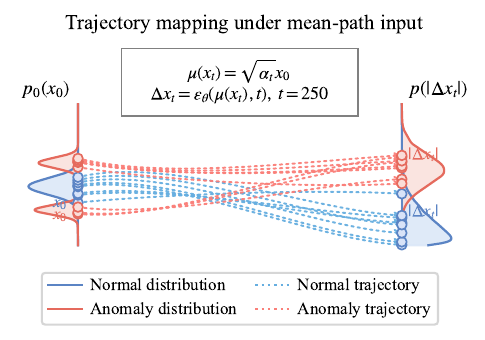}  
		\caption{}  
		\label{fig:2b}
	\end{subfigure}
	\vspace{-1mm}
	\caption{Trajectories of the one-step transition from the input data space $p(\boldsymbol{x}_0)$ to the output IRF space $p(|\Delta \boldsymbol{x}_t|)$. (a) Trajectory mapping under noisy-state input $\boldsymbol{x}_t$. (b) Trajectory mapping under mean-path input $\mu(\boldsymbol{x}_t)$.}
	\label{fig:trajectory}
	\vspace{-2mm}
\end{figure}

\subsection{A Toy Data Analysis on OSD-IRF}
\label{sec:toydata}
We construct one-dimensional toy data to illustrate how our proposed OSD-IRF works in uIAD. Normal data is generated from a Gaussian distribution while abnormal data is generated from a mixture Gaussian distribution, with probability density functions (PDFs) given by:
\begin{align}
	 & p_{\rm{normal}}(x) = \mathcal{N}(2.5,0.35^2), \quad p_{\rm{abnormal}}(x) = 0.5 \mathcal{N}(1.5,0.2^2)+ 0.5 \mathcal{N}(3.5,0.22^2)
\end{align}
During training, we generate 10,000 samples from $p_{\rm{normal}}(x)$ to train a DDPM, obtaining a well-trained noise function $\boldsymbol{\epsilon}{\boldsymbol{\theta}}(\cdot, t)$ parameterized by $\boldsymbol{\theta}$.
During inference, we set time step $t=250$ and we separately generate 6,000 normal samples $\boldsymbol{x}_{{\rm{normal}},t}^{1:6000}$ from $p_{\rm{normal}}(x)$ and 6,000 abnormal samples $\boldsymbol{x}_{{\rm{abnormal}},t}^{1:6000}$ from $p_{\rm{abnormal}}(x)$.
For each test sample in $\boldsymbol{X}_t=\{\boldsymbol{x}_{{\rm{normal}},t}^{1:6000}, \boldsymbol{x}_{{\rm{abnormal}},t}^{1:6000}\}$, we plot the one-step transition trajectories from the input data space $\boldsymbol{x}_0$ to the corresponding IRFs under both noisy-state input $\boldsymbol{x}_t$ and mean-path input $\mu({\boldsymbol{x}}_t)$. The results are illustrated in Fig. \ref{fig:trajectory}.
As can be seen, normal and abnormal patterns are clearly distinguishable in the IRF space derived from the mean-path input $\mu(\boldsymbol{x}_t)$ (See Fig. \ref{fig:trajectory} (b)), supporting the effectiveness of our model. Additionally, the one-step transition from input data to the corresponding IRF shown in Fig. \ref{fig:trajectory} demonstrates the inference efficiency of our proposed OSD-IRF. Importantly, comparing Fig. \ref{fig:trajectory} (a) and (b), the IRF under $\mu(\boldsymbol{x}_t)$ is clearly more robust than its counterpart under $\boldsymbol{x}_t$ for the uIAD task.
Therefore, we adopt the IRF under the mean-path input $\mu(\boldsymbol{x}_t)$ for the remainder of this paper.

\section{Experiments}

\subsection{Datasets}
We evaluate the effectiveness and efficiency of our model on three widely used uIAD benchmarks including MVTec-AD \cite{bergmann2019mvtec}, ViSA \cite{zou2022spot}, and MPDD \cite{jezek2021deep}.
MVTec-AD contains 15 classes, with a total of 2,629 normal images for training and 467 normal images plus 1,258 abnormal images for testing. The image resolution is $1024 \times 1024$. ViSA covers 12 classes, comprising 8,659 normal images for training, and 962 normal images with 1,200 abnormal images for testing. The image resolution is $256 \times 256$. MPDD has 6 classes, including 888 normal images for training and 176 normal images with 282 abnormal images for testing.

\subsection{Evaluation Metrics}
Following \cite{feng2025omiad}, we evaluate image/pixel-level AD performance using six metrics ($\%$) including Area Under the Receiver Operating Characteristic Curve (AU-ROC), Average Precision (AP), F1-score max (F1-max), and Area Under the Per-Region-Overlap (AU-PRO). We employ the Frames Per Second (FPS) to study the inference speed.

\subsection{Baseline AD Methods}
\label{sec:baseline}
We compare our model with most recently proposed uIAD methods, including UniAD \cite{you2022unified} and HVQTrans \cite{lu2023hierarchical}, SimpleNet \cite{liu2023simplenet}, DeSTSeg \cite{zhang2023destseg}, RD4AD \cite{deng2022anomaly}, DiAD \cite{he2024diffusion}, OmiAD \cite{feng2025omiad}, MDM \cite{feng2025omiad}, and InvAD \cite{sakai2025invad}. Among them, UniAD and HVQTrans are transformer-based reconstruction methods. SimpleNet, DeSTSeg, and RD4AD are CNN-based reconstruction methods. DiAD, OmiAD, and MDM are diffusion-based reconstruction methods. InvAD is diffusion-based density estimation method.

\subsection{Implementation Details}
For fair comparison, we evaluate our model on all datasets under a unified protocol following \cite{feng2025omiad}. An EfficientNet-B4 \cite{tan2019efficientnet} pre-trained on ImageNet-1k \cite{deng2009imagenet} is employed to extract image features. Based on the image features, a diffusion transformer (DiT) \cite{peebles2023scalable} is trained from scratch. The training follows a standard DDPM setting \cite{ho2020denoising} with a linear noise schedule and $T=1000$ time steps. AdamW \cite{loshchilov2017decoupled} is adopted as optimizer, and the model is trained for 300 epochs with a batch size equals to 32 for all datasets. During inference, the time step is empirically fixed to $t=500$ for all experiments.
All experiments are conducted on a single NVIDIA RTX 5880 Ada GPU (48G).

\begin{table}[t]
	\centering
	\caption{Quantitative results on different AD datasets under the multi-class setting. The best and the second-best results are highlighted in bold and underlined, respectively. For each row, the mAD averages the evaluation over all image- and pixel-level metrics (i.e., the 3rd to the 9th column). FPS measures the detection efficiency.}
	\vspace{2mm}
	\label{tab:iadresults}
	\setlength{\tabcolsep}{3.6pt}  
	\begin{tabular}{l l c c c c c c c c c}
		\toprule
		\multirow{2}{*}{\textbf{Dataset}} & \multirow{2}{*}{\textbf{Method}} & \multicolumn{3}{c}{\textbf{Image-level}} & \multicolumn{4}{c}{\textbf{Pixel-level}} & \multirow{2}{*}{\textbf{mAD}} & \multirow{2}{*}{\textbf{FPS}}                                                                                                 \\
		\cmidrule(lr){3-5} \cmidrule(lr){6-9}
		                                  &                                  & AU-ROC                                   & AP                                       & F1$_{\text{max}}$             & AU-ROC                        & AP               & F1$_{\text{max}}$ & AU-PRO           &                  &                  \\
		\midrule
		\multirow{9}{*}{MVTec-AD}
		                                  & RD4AD [CVPR'22]                  & 94.6                                     & 96.5                                     & 95.2                          & 96.1                          & 48.6             & 53.8              & 91.1             & 82.3             & 4.8              \\
		                                  & UniAD [NeurIPS'22]               & 96.5                                     & 98.8                                     & 96.2                          & 96.8                          & 43.4             & 49.5              & 90.7             & 81.7             & 5.3              \\
		                                  & SimpleNet [CVPR'23]              & 95.3                                     & 98.4                                     & 95.8                          & 96.9                          & 45.9             & 49.7              & 86.5             & 81.2             & N/A              \\
		                                  & DeSTSeg [CVPR'23]                & 89.2                                     & 95.5                                     & 91.6                          & 93.1                          & \textbf{54.3}    & 50.9              & 64.8             & 77.1             & N/A              \\
		                                  & DiAD [AAAI'24]                   & 97.2                                     & 99.0                                     & 96.5                          & 96.8                          & \underline{52.6} & 55.5              & 90.7             & 84.0             & 0.1              \\
		                                  & HVQ-Trans [NeurIPS'23]           & 98.0                                     & 99.5                                     & 97.5                          & 97.3                          & 48.2             & 53.3              & 91.4             & 83.6             & 5.6              \\
		                                  & MDM [ICML'25]                    & 87.2                                     & 94.2                                     & 91.0                          & 94.8                          & 39.5             & 44.8              & 88.0             & 77.1             & 1.9              \\
		                                  & OmiAD [ICML'25]                  & \underline{98.8}                         & \textbf{99.7}                            & \textbf{98.5}                 & \textbf{97.7}                 & \underline{52.6} & \underline{56.7}  & \underline{93.2} & \underline{85.3} & 39.4             \\
		                                  & {InvAD [CVPR'26]}                & \textbf{99.0}                            & \underline{99.6}                         & \textbf{98.5}                 & \underline{97.5}              & 46.5             & 52.3              & 92.7             & 83.7             & \underline{88.1} \\
		                                  & \textbf{OSD-IRF (Ours)}          & \textbf{99.0}                            & \underline{99.6}                         & \underline{98.4}              & \textbf{97.7}                 & \textbf{54.3}    & \textbf{57.4}     & \textbf{93.5}    & \textbf{85.7}    & \textbf{133}     \\
		\midrule
		\multirow{8}{*}{ViSA}
		                                  & RD4AD [CVPR'22]                  & 92.4                                     & 92.4                                     & 89.6                          & 98.1                          & 38.0             & 42.6              & 91.8             & 77.8             & 4.9              \\
		                                  & UniAD [NeurIPS'22]               & 88.8                                     & 90.8                                     & 85.8                          & 98.3                          & 33.7             & 39.0              & 85.5             & 74.6             & 4.6              \\
		                                  & SimpleNet [CVPR'23]              & 87.2                                     & 87.0                                     & 81.8                          & 96.8                          & 34.7             & 37.8              & 81.4             & 72.4             & N/A              \\
		                                  & DeSTSeg [CVPR'23]                & 88.9                                     & 89.0                                     & 85.2                          & 96.1                          & 39.6             & 43.4              & 67.4             & 72.8             & N/A              \\
		                                  & DiAD [AAAI'24]                   & 86.8                                     & 88.3                                     & 85.1                          & 96.0                          & 26.1             & 33.0              & 75.2             & 70.1             & 0.1              \\
		                                  & HVQ-Trans [NeurIPS'23]           & 93.2                                     & 92.8                                     & 87.6                          & 98.7                          & 35.0             & 39.6              & 86.3             & 76.2             & 5.0              \\
		                                  & OmiAD [ICML'25]                  & 95.3                                     & 96.0                                     & 91.2                          & 98.9                          & \underline{40.4} & \underline{44.1}  & 89.2             & 79.3             & 35.3             \\
		                                  & {InvAD [CVPR'26]}                & \textbf{96.9}                            & \underline{97.2}                         & \underline{93.7}              & \textbf{99.1}                 & 39.2             & 43.1              & \underline{92.9} & \underline{80.3} & \underline{74.1} \\
		                                  & \textbf{OSD-IRF (Ours)}          & \underline{96.8}                         & \textbf{97.2}                            & \textbf{93.9}                 & \underline{99.0}              & \textbf{42.6}    & \textbf{45.9}     & \textbf{93.5}    & \textbf{81.3}    & \textbf{105.9}   \\
		\midrule
		\multirow{8}{*}{MPDD}
		                                  & RD4AD [CVPR'22]                  & 84.1                                     & 83.2                                     & 84.1                          & 98.1                          & 35.2             & 38.7              & 93.4             & 73.8             & 4.7              \\
		                                  & UniAD [NeurIPS'22]               & 82.2                                     & 87.1                                     & 85.1                          & 95.1                          & 18.9             & 25.0              & 81.9             & 67.9             & 5.8              \\
		                                  & SimpleNet [CVPR'23]              & 90.6                                     & 94.1                                     & 89.7                          & 97.1                          & 33.6             & 35.7              & 90.0             & 75.8             & N/A              \\
		                                  & DeSTSeg [CVPR'23]                & 93.0                                     & 95.1                                     & 90.6                          & 94.1                          & 33.2             & 37.6              & 59.8             & 71.9             & N/A              \\
		                                  & DiAD [AAAI'24]                   & 74.6                                     & 82.1                                     & 82.5                          & 93.0                          & 15.9             & 21.2              & 78.4             & 64.0             & 0.1              \\
		                                  & HVQ-Trans [NeurIPS'23]           & 86.5                                     & 88.1                                     & 85.8                          & 96.7                          & 27.6             & 31.4              & 86.9             & 71.9             & 6.2              \\
		                                  & OmiAD [ICML'25]                  & 93.7                                     & 95.5                                     & 90.9                          & \underline{98.6}              & 37.6             & 42.3              & \underline{94.0} & 78.9             & 49.8             \\
		                                  & InvAD [CVPR'26]                  & \underline{96.5}                         & \underline{96.5}                         & \underline{94.4}              & 98.3                          & \underline{40.1} & \underline{43.9}  & 91.2             & \underline{80.1} & \underline{120}  \\
		                                  & \textbf{OSD-IRF (Ours)}          & \textbf{98.3}                            & \textbf{98.5}                            & \textbf{96.3}                 & \textbf{98.9}                 & \textbf{48.4}    & \textbf{50.0}     & \textbf{95.0}    & \textbf{83.8}    & \textbf{212.4}   \\
		\bottomrule
	\end{tabular}
	\vspace{-2mm}
\end{table}

\subsection{Comparison with SOTA AD Methods}
We compare our proposed OSD-IRF against prior uIAD baseline methods introduced in Sec. \ref{sec:baseline} and summarize the results in Table \ref{tab:iadresults}. Our model generally achieves SOTA performance on all three datasets in terms of mAD, which averages the evaluation across all the six AD metrics. We attribute this to the fact that our defined IRF space is more robust and better suited for the uIAD task. Furthermore, OSD-IRF is faster at test time than all competitors, an appealing property for real-world uIAD applications. For example, our model achieves nearly 2 $\times$ inference speedup compared with InvAD. This advantage stems from the one-step inference property of the IRF, which does not rely on computations over multiple time steps, as shown in Eq. \ref{eq:irf2}.
The most relevant works to ours are OmiAD and InvAD, both of which develop diffusion-based methods for fast and accurate uIAD. OmiAD performs anomaly detection in the input data space $\boldsymbol{x}_0$ and uses adversarial distillation learning (ADL) to reduce the denoising process to a single step (NFE = 1). Although effective, we find that ADL is difficult to stabilize. InvAD, on the other hand, performs AD in the noisy latent space $\boldsymbol{x}_T$ and uses inverse DDIM to reduce the diffusion process to two or three steps. In contrast, our OSD-IRF explores a novel IRF space $\Delta \boldsymbol{x}_t$ for anomaly discovery, which is shown to be more robust than OmiAD's $\boldsymbol{x}_0$ and InvAD's $\boldsymbol{x}_T$, as evidenced by the results in Table \ref{tab:iadresults}.

\subsection{The Advantages of IRF in AD Task}
\label{sec:benefits}
To better understand the strengths of our model, we clarify the benefits of using the IRF for AD as follows. First, compared to conditional diffusion models such as DiAD and OmiAD, our model requires no carefully designed conditioning. Second, compared to methods that rely on intricate training paradigms (e.g., OmiAD), our model is nearly training-free, as it only requires training a standard DDPM. Third, compared to almost training-free models such as InvAD, our model achieves higher AD accuracy and faster inference.
We attribute all of these advantages to the desirable properties of our proposed IRF space, namely that it is both distinguishable for normal versus abnormal patterns and independent of multi-step inference.


\begin{figure}[t]
	\centering
	\centerline{\includegraphics[width=16cm]{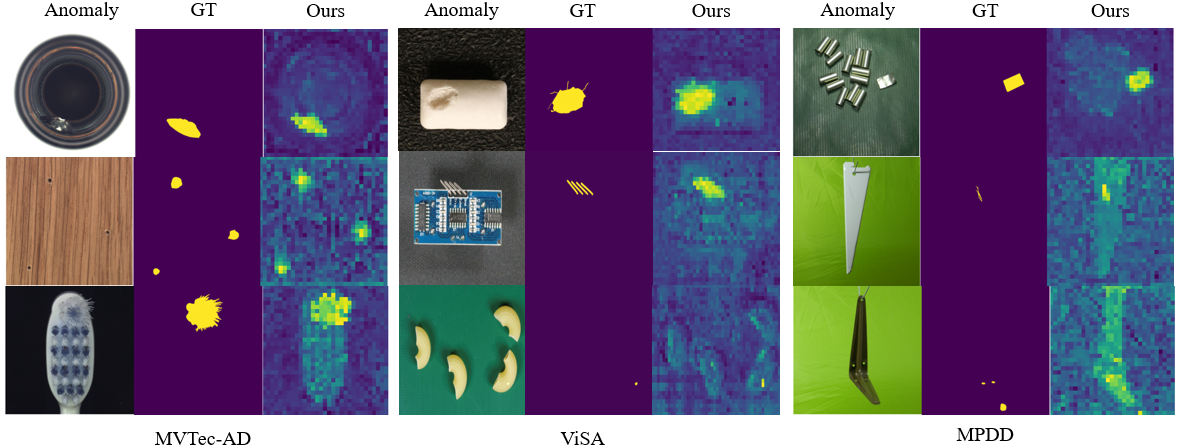}}
	\vspace{-2mm}
	\caption{Visualizations of anomaly localization by our model on MVTec‑AD, ViSA, and MPDD.}
	\label{fig:visualization}
	\vspace{-2mm}
\end{figure}

\subsection{Visualization Results}
We present qualitative results in Fig. \ref{fig:visualization} to visualize the anomaly localization performance of our model on three datasets including MVTec-AD, ViSA, and MPDD. As we can see that OSD-IRF effectively distinguish anomalies in the IRF space, accurately recognizing abnormal patterns and then localizing anomalous regions.

\section{Conclusion}
In this paper, we proposed OSD-IRF, a one-step diffusion framework for unsupervised industrial anomaly detection (uIAD) based on inverse residual fields (IRF). Unlike previous reconstruction-based or density estimation-based diffusion models that operate in the input data space or noisy latent space, OSD‑IRF performs anomaly detection in a novel IRF space, defined as the predicted noise between two adjacent time steps. Our key observation that normal and abnormal patterns become clearly distinguishable in this space has rarely been reported in prior work. By leveraging the property that the IRF holds for any neighboring time step, our method requires only a single diffusion step, significantly reducing inference latency.
We demonstrated that our model is both effective and efficient. Extensive experiments on three widely used uIAD benchmarks (MVTec-AD, ViSA, and MPDD) showed that our model achieves SOTA or competitive performance across six metrics, with roughly a 2$\times$ inference speedup without any distillation. Moreover, OSD-IRF is plug-and-play: it can be applied to any pre-trained diffusion model regardless of architecture or training objective.
We believe the concept of inverse residual fields may inspire new directions in fast density estimation for real-time industrial inspection and broader anomaly detection tasks.

\bibliographystyle{unsrtnat}
\bibliography{references}  

\end{document}